\documentclass[runningheads]{llncs}

\usepackage{eccv}

\usepackage{eccvabbrv}

\usepackage{graphicx}
\usepackage{booktabs}
\usepackage[accsupp]{axessibility}  % Improves PDF readability for those with disabilities.
\usepackage{amsmath}
\usepackage{amssymb}
\usepackage{makecell}
\usepackage{subcaption}
\usepackage{caption}
\usepackage{multirow}
\usepackage{multicol}
\usepackage{soul}
\usepackage{comment}
\usepackage{bbding}
\usepackage{xspace}
\makeatletter
\DeclareRobustCommand\onedot{\futurelet\@let@token\@onedot}
\def\@onedot{\ifx\@let@token.\else.\null\fi\xspace}

\def\eg{\emph{e.g}\onedot} 
\def\ie{\emph{i.e}\onedot}

\def\wrt{w.r.t\onedot}

\makeatother

\usepackage[dvipsnames]{xcolor}

\usepackage{hyperref}

\usepackage{orcidlink}

\begin{document}

\title{RealViformer: Investigating Attention for Real-World Video Super-Resolution} 

\titlerunning{RealViformer}

\author{Yuehan Zhang\orcidlink{0000-0002-5017-0097} \and
Angela Yao\orcidlink{0000-0001-7418-6141}}

\authorrunning{Y.~Zhang and A.~Yao}

\institute{National University of Singapore\\
\email{\{zyuehan,ayao\}@comp.nus.edu.sg}}

\maketitle
\begin{abstract}
In real-world video super-resolution (VSR), videos suffer from in-the-wild degradations and artifacts. VSR methods, especially recurrent ones, tend to propagate artifacts over time in the real-world setting and are more vulnerable than image super-resolution. This paper investigates the influence of artifacts on commonly used covariance-based attention mechanisms in VSR. Comparing the widely-used spatial attention, which computes covariance over space, versus the channel attention, we observe that the latter is less sensitive to artifacts.
However, channel attention leads to feature redundancy, as evidenced by the higher covariance among output channels. As such, we explore simple techniques such as the squeeze-excite mechanism and covariance-based rescaling to counter the effects of high channel covariance. Based on our findings, we propose RealViformer. This channel-attention-based real-world VSR framework surpasses state-of-the-art on two real-world VSR datasets with fewer parameters and faster runtimes. The source code is available at \url{https://github.com/Yuehan717/RealViformer}.
\end{abstract}    
\section{Introduction}
\label{sec:intro}

\begin{figure}
\centering
\begin{minipage}[b]{0.40\textwidth}
\centering
\begin{subfigure}[t]{\textwidth}
    \centering
        \includegraphics[width=0.85\linewidth]{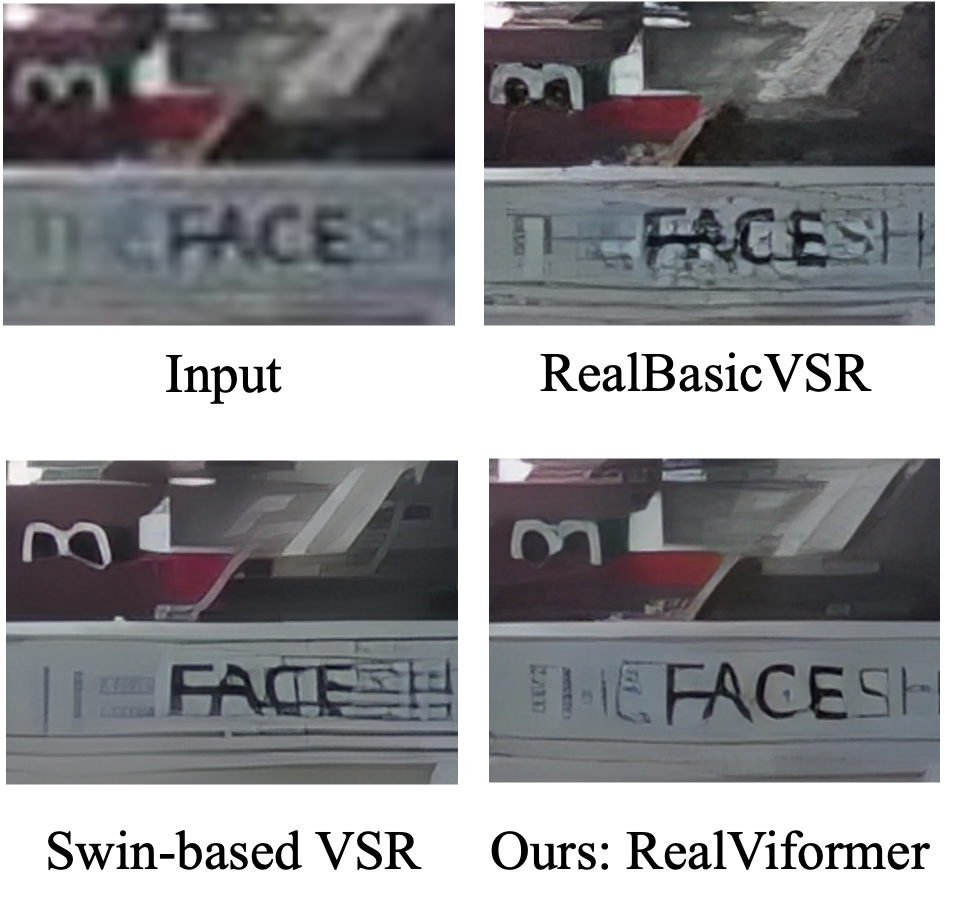}
        \subcaption{Visual comparisons.}
    \label{fig:teaser}
\end{subfigure}
\end{minipage}
\begin{minipage}[b]{0.45\textwidth}
\begin{subfigure}[t]{\textwidth}
    \centering
        \includegraphics[width=\linewidth]{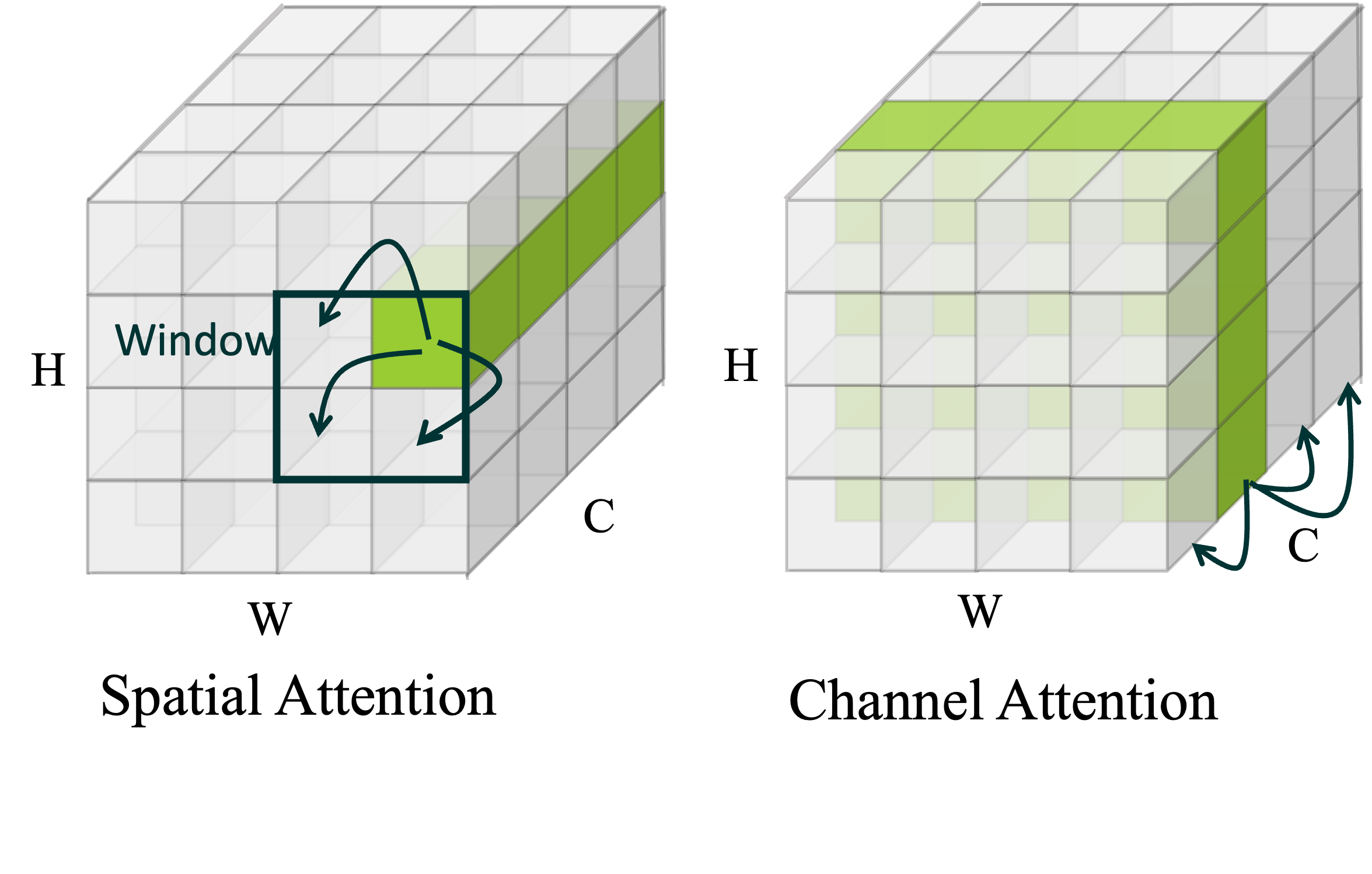}
        \subcaption{Schematics of two attentions.}
    \label{fig:sche_attn}
\end{subfigure}
   
\end{minipage}
\caption{(a) Designing a RWVSR transformer is not trivial. A Swin-based transformer suited for standard VSR hallucinates more lines than a RealBasicVSR, a convolutional state-of-the-art. We propose RealViformer based on our investigation of attention under the RWVSR setting. RealViformer generates details with fewer artifacts than RealBasicVSR~\cite{chan2021basicvsr} and the Swin-based VSR model. (b) Schematic for spatial and channel attention. Spatial attention aggregates features based on pixel representations. Channel attention takes $H\times W$ feature map for matching across channels.}
\end{figure}

Video super-resolution (VSR) recovers a high-resolution (HR) sequence of frames from its low-resolution (LR) counterpart. Recurrent convolutional approaches are commonly used in VSR with standard settings, assuming the LR frames are downsampled from HR frames with known kernels.~\cite{chan2021basicvsr, fuoli2019efficient, isobe2020video}. However, in real-world VSR (RWVSR), the low-resolution videos are not simply downsampled versions of their high-resolution counterparts. Instead, they feature complex degradations that arise from the camera imaging system, compression, internet transmission, and other factors. These degradations make architecture design for RWVSR challenging, as artifacts and degradations tend to propagate and get exaggerated over the recurrent connection~\cite{chan2022investigating,xie2022mitigating}.

Recently, transformer architectures have replaced convolutional architectures as state-of-the-art for standard VSR. While attention mechanisms have replaced convolution operations, most methods retain the recurrent connection to aggregate the information over time~\cite{shi2022rethinking, liang2022recurrent}. Yet, such architectures do not always perform well on RWVSR. For instance, a Swin-based model~\cite{liang2021swinir} designed for standard VSR, when applied to a real-world input frame, as shown in \cref{fig:teaser}, generates more artificial lines than the convolutional model RealBasicVSR~\cite{chan2022investigating}.

Why should transformers perform well on (synthetic) standard VSR but poorly in real-world cases? We speculate that standard VSR transformers benefit from the similarity-based matching of the attention mechanism, which accurately aggregates information spatially and temporally~\cite{liang2021swinir, shi2022rethinking, liang2022recurrent}. However, when input degradation exists, the aggregated information becomes \textit{less reliable} because the attention queries may be derived from both true source video and artifacts.  

This work investigates and sheds light on the sensitivity of attention in real-world settings. We compare two covariance-based attention mechanisms used in low-level Transformers: spatial attention~\cite{liang2021swinir, shi2022rethinking} and channel attention~\cite{zamir2022restormer}. Spatial attention takes pixel-wise features as keys and queries and estimates their covariances across spatial positions. The most popular form is window-based attention~\cite{chen2021pre}; The shift-window scheme from Swin Transformers~\cite{liu2021swin} enables the model to access 
distant spatial ranges~\cite{shi2022rethinking, cao2021video, liang2021swinir} without computational blow-up. Spatial attention is widely used for video and image super-resolution, albeit in the standard setting.  
Channel attention~\cite{zamir2022restormer} estimates covariances across channels (see \cref{fig:sche_attn}) and collapses the spatial extent of a feature map. It defines the number of queries and keys by channel numbers rather than spatial resolution. Although established for deblurring or denoising~\cite{zamir2022restormer}, channel attention's efficacy in super-resolution remains to be determined.

Our experiments show that channel attention is less sensitive to artifacts than the spatial counterparts, resulting in higher performance gains in RWVSR. However, it is also revealed that channel attention leads to feature channels with higher covariance. From a learning perspective, a high covariance is undesirable because it is a strong indicator of feature redundancy~\cite{bardes2021vicreg,cogswell2015reducing, hua2021feature}. Therefore, using channel attention naively will have limited improvements over existing RWVSR state-of-the-art methods. Such a finding has wide-reaching impacts as channel attention is used increasingly for low-level vision.

To verify our findings, we explore established mechanisms to counter the effects of feature redundancy - simple techniques such as squeeze-excite and covariance-based rescaling improve the vanilla channel attention design. From these outcomes, we propose RealViformer, a new transformer real-world VSR model. 
RealViformer performs channel attention between the current frame feature and the propagated hidden state to limit model-produced artifacts. The model then reconstructs features through improved channel attention modules featuring squeeze-and-excite and covariance-based channel rescaling mechanisms. With our effective designs, RealViformer achieves state-of-the-art performance with fewer parameters on challenging synthetic video datasets and two real-world video datasets collected from different scenes.  

Summarizing our contributions in order of importance, our paper
\begin{itemize}
    \item investigates the differences between spatial and channel attention for RWVSR. Spatial attention, although widely used, is revealed to be highly sensitive to the noise and degradations common in RWVSR sequences, while channel attention is more robust.
    \item reveals that naively applying channel attention increases channel covariance, which is problematic for learning; this overlooked fact has a wide-reaching impact as channel attention becomes more used in low-level vision.
    \item empirically verifies the negative effect of high channel covariance by countering it with established techniques, based on which we develop the RealViformer for RWVSR. Our simple modification surpasses state-of-the-art despite using less compute.
\end{itemize}
\section{Related Work}

\textbf{Standard Video Super-Resolution}
models focus on architecture design to use temporal information better~\cite{chan2022basicvsr++,isobe2020video, chan2021basicvsr, cao2021video}. Previous research starts from slide-window-based~\cite{wang2019edvr,isobe2020video} to recurrent-based frameworks~\cite{chan2021basicvsr, chan2022basicvsr++} for using distant-frame information. Recent works introduce Transformer blocks into existing recurrent frameworks to overcome the locality limitation of convolution and accurately match abundant information for feature reconstruction~\cite{shi2022rethinking, liang2022vrt, liang2022recurrent}.

\noindent\textbf{Real-world video super-resolution} focuses on modeling, removing, and limiting the impact of real-world degradations. Existing works are convolutional models and focus on designing losses or modules for degradation processing.
DBVSR~\cite{pan2021deep} explicitly estimates the degradation kernel through a sub-network. 
RealBasicVSR~\cite{chan2022investigating} tries to `clean' artifacts through a processing module for BasicVSR~\cite{chan2021basicvsr}. In a similar approach, FastRealVSR~\cite{xie2022mitigating} borrows an external pool of blur and sharpening filters to `clean' the hidden states. Other recent works~\cite{jeelani2023expanding,song2024negvsr} advance the synthesis method for paired training data. Instead, we focus on investigating the function of attention in RWVSR rather than architecture design or data synthesis.

\noindent\textbf{Attention mechanisms} have been widely applied for low-level vision tasks~\cite{dai2019second,zhang2018image,mei2021image, zhang2018image}. Transformers with covariance-based attention are the most prevalent~\cite{zamir2022restormer,liang2021swinir, shi2022rethinking} for standard VSR. Most existing methods adopt shift-window-based spatial attention~\cite{liu2021swin} to aggregate information from other positions within or across frames. In contrast, Restormer~\cite{zamir2022restormer} computes the covariance among channels and shows its effectiveness for multiple image restoration tasks. More recent works stitch spatial and channel attention together to enlarge the receptive field~\cite{wang2023omni,chen2023activating} for standard image super-resolution. Instead, we investigate attention mechanisms in terms of their sensitivity to real-world degradation for the first time and develop an effective real-world VSR Transformer. 
\section{Explorations on Attention}
\label{sec:exploration}
\cref{sec:attndefs} defines the VSR task and the two attention mechanisms. \cref{sec:investigation} compares the channel and spatial attentions' sensitivity to query artifacts and effects on real-world VSR performance. \cref{sec:limitation} reveals that channel attention leads to higher covariance among channels and explores mitigating options.

\subsection{Preliminaries}
\label{sec:attndefs}
\noindent\textbf{Standard vs. Real-World VSR.}
Given a low-resolution (LR) video sequence with $T$ frames $I^{L}\in\mathcal{R}^{T\times\!H\times\!W\times K}$, VSR models reconstruct a high-resolution sequence $I^{H}\in\mathcal{R}^{T\times\!sH\times sW\times K}$, where $H\!\times\!W$ is the input spatial resolution, $K$ is the number of input channels and $s$ is the scaling factor. In the standard setting of VSR, $I^L_t$, where $t\in\{0,...,T\}$, is assumed as the downsampled version of $I^H_t$, defined as: $I^L_t = (I^H_t)\downarrow_{\frac{1}{s}}$. Both training and testing datasets follow this formulation to generate LR-HR pairs given $I^H$. In a real-world setting, there is no closed formulation for the relationship between $I^L_t$ and $I^H_t$ due to the unknown distribution of real-world degradations. Real-ESRGAN~\cite{wang2021real} proposed a widely-used training setting that randomly applies synthesized blur, noise, compression, and resizing to the HR frames to generate paired LR frames with complex degradations. The testing datasets are either synthesized by the same pipeline in training or collected from diverse real-world sources~\cite{chan2022investigating,yang2021real}. The synthesized testing sets have paired ground-truth sequences and are evaluated by full-reference metrics, \eg PSNR and LPIPS~\cite{zhang2018unreasonable}; real-world datasets are always without ground truth and require no-reference metrics, such as NRQM~\cite{ma2017learning}.

\noindent\textbf{Attention Definitions.}
In Transformers, the attention modules project layer-normalized tensor $X\in \mathbb{R}^{C\times H\times W}$ to query $Q$, and tensor $Y\in \mathbb{R}^{\hat{C}\times \hat{H}\times \hat{W}}$ to key $K$ and value $V$, where $H\!\times\!W$ and $\hat{H}\!\times\!\hat{W}$ are the spatial resolution of the normalized tensors.\footnote{Note we define mutual attention as the default; self-attention is a special form of mutual attention where $Y\!=\!X$.} 
The attention map $\mathcal{A}$ is generated by calculating the covariance between $Q$ and $K$, followed by a softmax function, before being applied to the value $V$ to produce output $O$.
Spatial and channel attention differ in tensor dimension taken for the covariance calculation.

Spatial attention generates an $\mathbb{R}^{HW\times \hat{H}\hat{W}}$ attention map, $\mathcal{A}_s$, by computing the covariance between features in the query ($Q_s$) and key ($K_s$) at each spatial position. 
The query, key and values are computed as $Q_{s}\!=\!W_{s}^{Q}X$, 
$K_{s}\!=\!W_{s}^{K}Y$, $V_{s}\!=\!W_{s}^{V}Y$, where $W_{s}^{Q} \in \mathbb{R}^{D_{s}\!\times\!C}$, $\{W_{s}^{K}, W_{s}^{V} \}\in \mathbb{R}^{D_{s}\!\times \!\hat{C}}$, and $D_{s}$ is the dimension of projections. The attention map $\mathcal{A}_{s}$ and output attention features $O_s$ are:
\begin{equation}
    \mathcal{A}_{s} = \text{softmax}(Q_s^{T}K_s/\sqrt{D_s}),\;
    O_s=\mathcal{A}_{s}V_s^{T},
    \label{eq:attention_s}
\end{equation}
where $Q_s$ ($K_s$) is reshaped to matrix of $\mathbb{R}^{D_{s}\times HW}$ ($\mathbb{R}^{D_{s}\times \hat{H}\hat{W}}$). We omit the later reshaping operation for brevity. $\mathcal{A}_{s}$ exhaustively relates every spatial location to every other location in space. A window-based implementation~\cite{liang2021swinir} limits the correlations to windows of $\omega\!\times\!\omega$, reducing the map size to $\mathbb{R}^{\omega^{2}\times\omega^2}$. 

Channel attention applies convolutions on $X$ to get query ($Q_c$) and $Y$ for key ($K_c$) and value ($V_c$)~\cite{zamir2022restormer}, after which features are the same in spatial resolution (assumed to be $H\times\!W$ in the following context).
It estimates covariance across channels to yield a {$C\times\hat{C}$} sized map $\mathcal{A}_c$ and output features $O_c$: 
\begin{equation}
    \mathcal{A}_{c} = \text{softmax}(Q_cK_c^{T}/\alpha),\;
    O_c=\mathcal{A}_cV_c,
    \label{eq:attention_c}
\end{equation} 
where $Q_c$ is reshaped to $\mathbb{R}^{C\times HW}$, $K_c$ and $V_c$ are reshaped to $\mathbb{R}^{\hat{C}\times HW}$ and $\alpha$ is learnable scaling parameter. Because channel attention computes correlations between features of size $\mathbb{R}^{1\times\!HW}$, it has a larger spatial context than the $\mathbb{R}^{D_s\times\!1}$-sized features in spatial attention (see~\cref{fig:sche_attn}).

\subsection{Attention in Real-World VSR}
\label{sec:investigation}
\noindent\textbf{Sensitivity to Query Artifacts.}
In standard VSR, attention is used to aggregate information temporally and spatially~\cite{shi2022rethinking, liang2022recurrent}. The assumption is that queries can match beneficial cues for super-resolution. But what if the queries themselves are unreliable? We speculate this to be true in real-world VSR, where inputs have artifacts and degradations. 

\begin{figure}[h]
    \centering
    \includegraphics[width=0.5\linewidth]{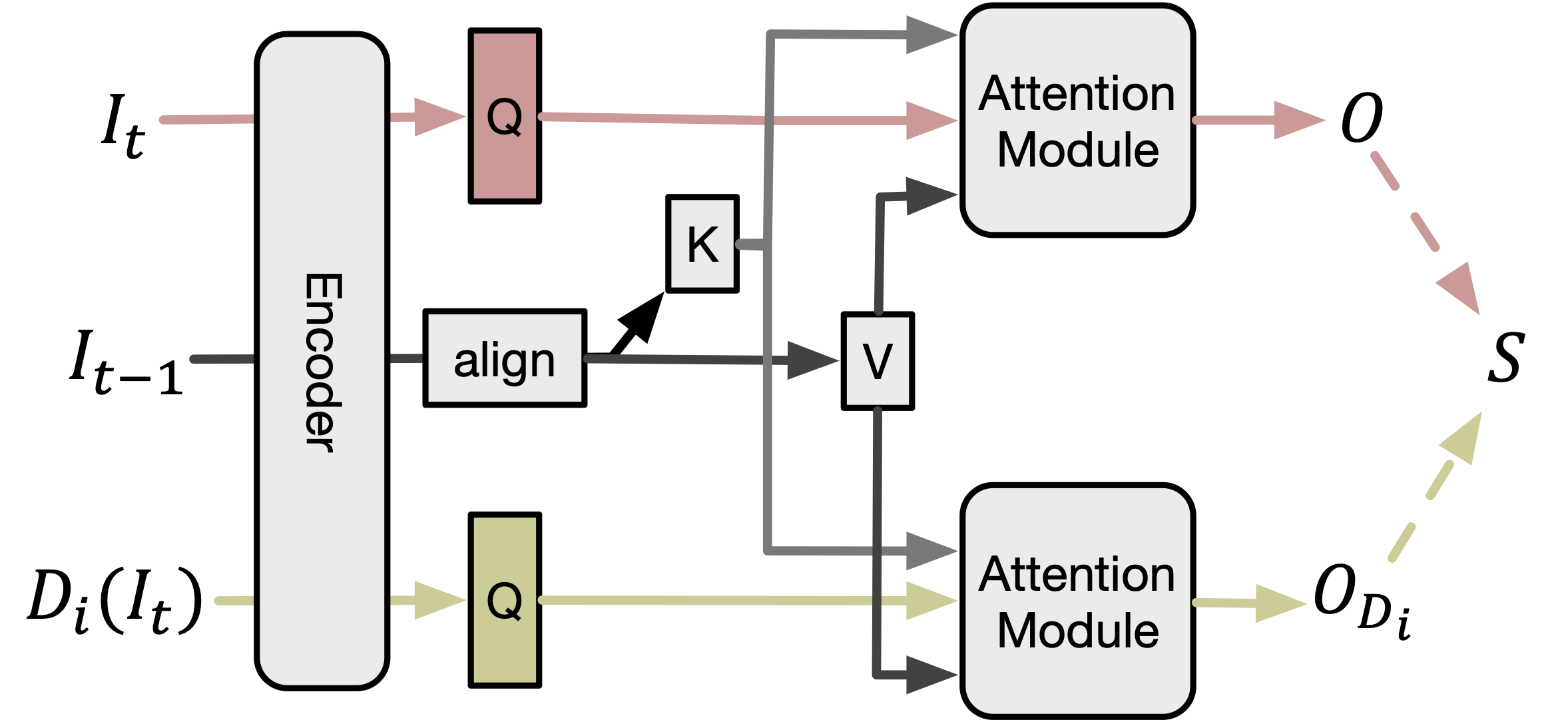}
    \caption{Schematic for sensitivity comparison. ${I_{t-1}, I_{t}}$ are downsampled but clean frames at times $t$ and $t-1$. $D_i(.)$ apply degradations to $I_t$, where $D_i \in \text{\{blur, noise, compression\}}$. $O$ and $O_{D_i}$ are output features of the attention module. Queries are from the embedding at time $t$, and keys and values are from time $t-1$. Higher cosine similarities $S$ between attention output features $O$ and $O_{D_i}$ reflect less sensitivity to artifacts in queries.}
    \label{fig:query_change}
\end{figure}

How affected are spatial and channel attention mechanisms by query artifacts?
\cref{fig:query_change} shows how we compare, using cosine similarity, the attention outputs based on queries from the same frame with and without certain degradations. Using encoding layers from a standard convolutional VSR model, we perform the attention operations defined in \cref{sec:attndefs} by taking embeddings of frame $I_t$ as query and $I_{t-1}$ as key and value. $I_t$ and $I_{t-1}$ are clean without degradation other than downsampling, and we represent the output feature of the attention module as $O$. When additional degradation $D_i$ is applied to $I_t$, we represent the corresponding output as $O_{D_i}$. A smaller deviation of $O_{D_i}$ from $O$ indicates lower sensitivity to artifacts in the query. 
\begin{table}[h]
    \centering
    \captionof{table}{Cosine similarity between $O$ and $O_{D_i}$, attention outputs without and with query degradation. Outputs of channel attention change less under query degradation.}
    \begin{tabular}{c|c|c|c}
          \multirow{2}{*}{Module} & \multicolumn{3}{c}{$D_i$}\\
           \cline{2-4}
           & blur & noise & compression\\
        \hline
        Spatial attention & 0.75 &0.92 &0.84 \\
        Channel attention& 0.98 &0.99 &0.99
    \end{tabular}
    \label{tab:S_attention}
\end{table}

\cref{tab:S_attention} shows the cosine similarity between $O$ and $O_{D_i}$ for spatial and channel attention modules.
Curiously, $O_{D_i}$ based on the channel attention module is more similar to the $O$ matched by the degradation-free query, indicating that channel attention is less sensitive to query artifacts.
Intuitively, the lower sensitivity of channel attention is related to the larger spatial context used for feature matching. Given a deep feature of size $\mathbb{R}^{C\times\!H\times\!W}$, channel attention uses feature sized $\mathbb{R}^{1\!\times\!HW}$ to calculate the covariance across channels. As such, feature aggregation is based on global information observed in a large normalized spatial context. 
Instead, the covariance of spatial attention is for features at each location sized $\mathbb{R}^{C\!\times\!1}$, so it is likely more sensitive to local value changes from artifacts.

\begin{figure}[h]
\centering
\begin{minipage}[h]{0.55\linewidth}
\centering
\subfloat[Recurrent Baseline]{
\includegraphics[width=0.70\textwidth]{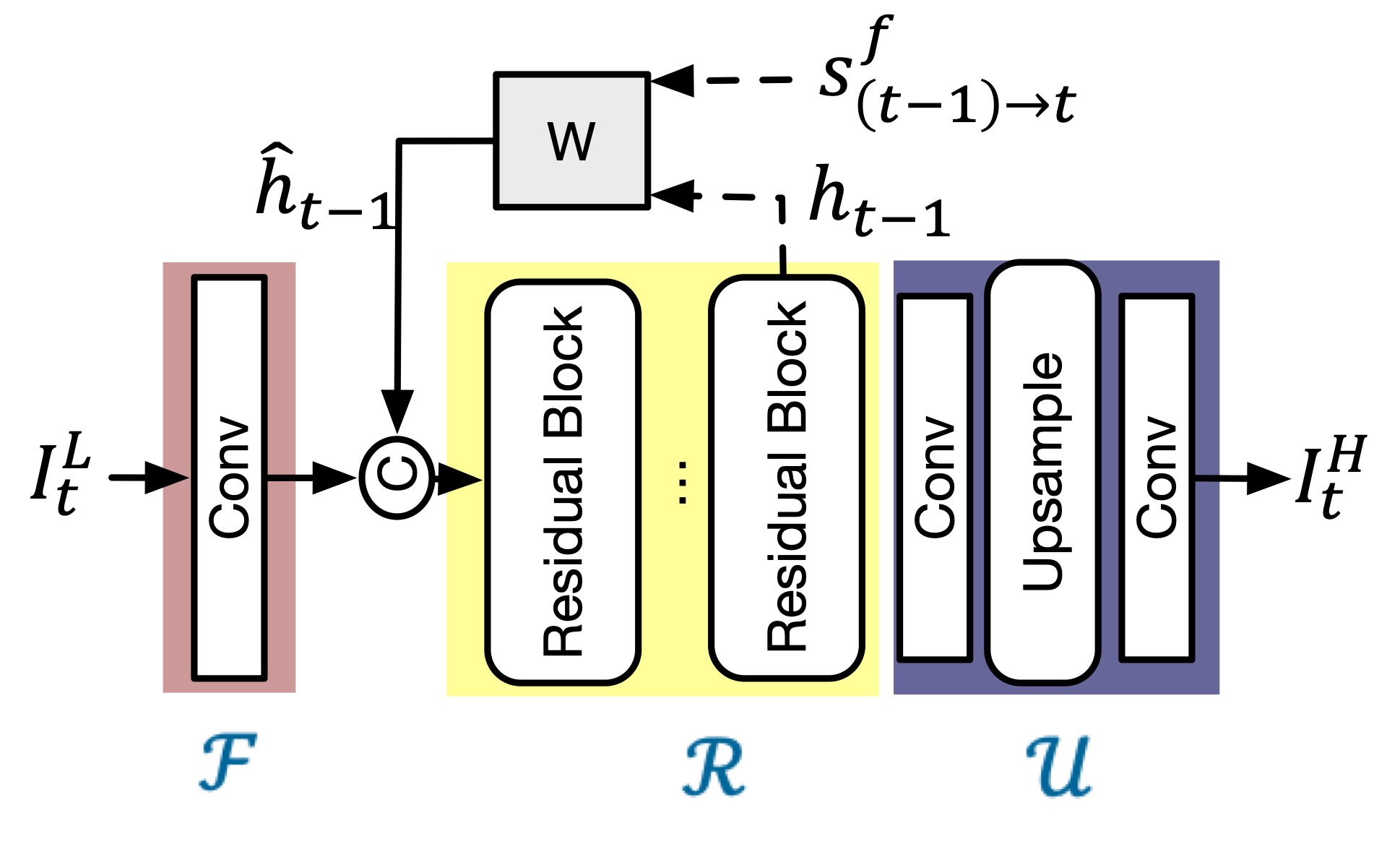}
\label{subfig:baseline}
}
\end{minipage}
\begin{minipage}[h]{0.35\linewidth}
\subfloat[Attention]{
\includegraphics[width=\linewidth]{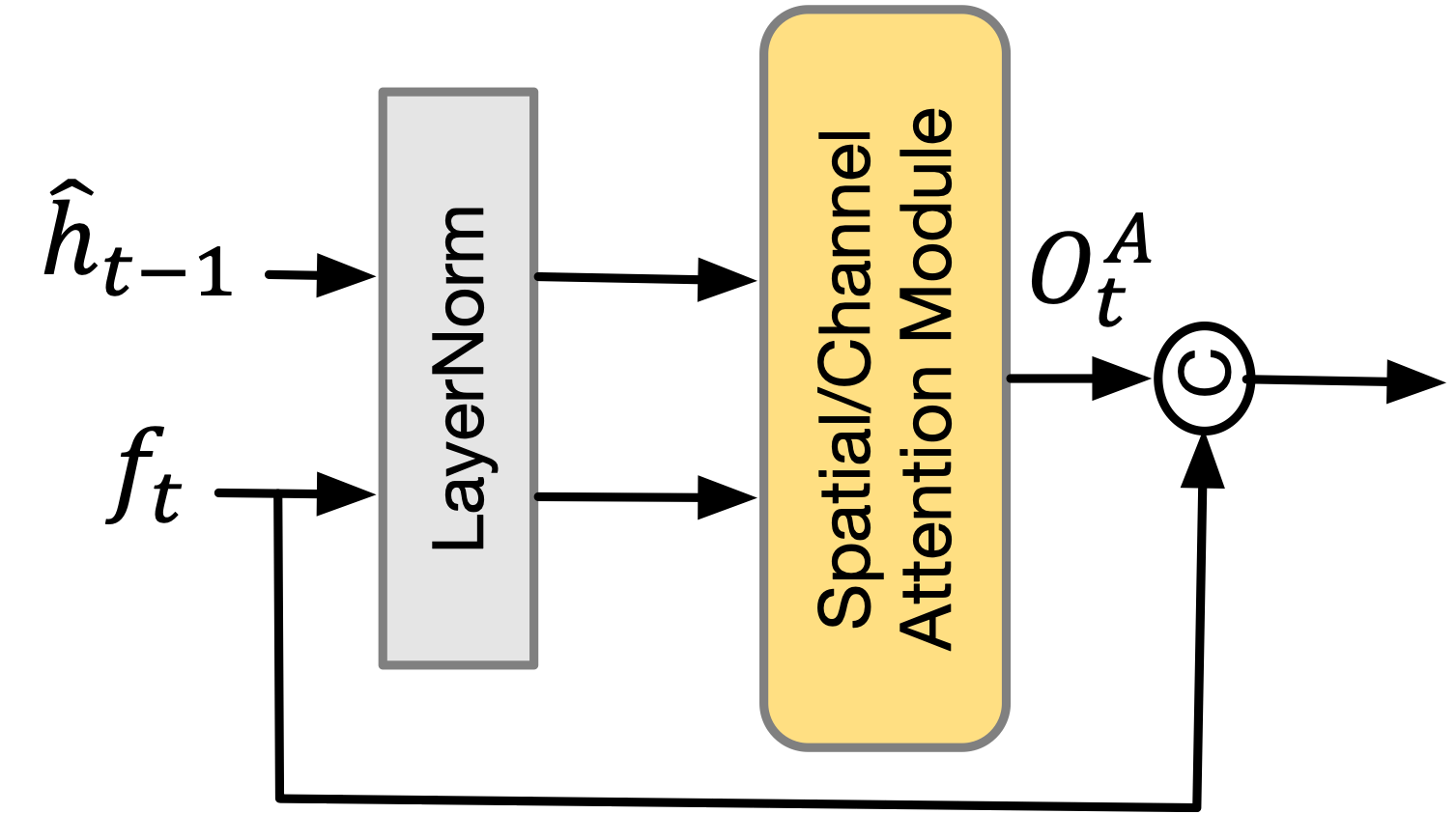}
\label{subfig:ch_merge}
}
\end{minipage}
\caption{(a) The recurrent baseline in \cref{sec:investigation} has a shallow mapping module $\mathcal{F}$, reconstruction module $\mathcal{R}$, upsampling module $\mathcal{U}$ and warping function $W$. $W$ aligns the hidden state $h_{t-1}$ to feature at $t$ based on optical flow $s^f_{(t-1)\rightarrow\!t}$. All residual blocks are convolutional. The concatenation between $f_t$ and $\hat{h}_{t-1}$ are replaced with the spatial or channel attention modules in (b) to compare the effect of attention. (b) The attention module first applies layer normalization to $f_{t}$ and $\hat{h}_{t-1}$ and then performs channel or spatial attention according to \cref{sec:attndefs}. The output feature $O^{A}_{t}$ concatenated with $f_t$ is processed by the module $\mathcal{R}$ in (a).}

\label{fig:sub1_backbone}
\end{figure}

\begin{figure*}[!t]
\begin{subfigure}[t]{0.32\textwidth}
    \centering 
    \includegraphics[width=\linewidth]{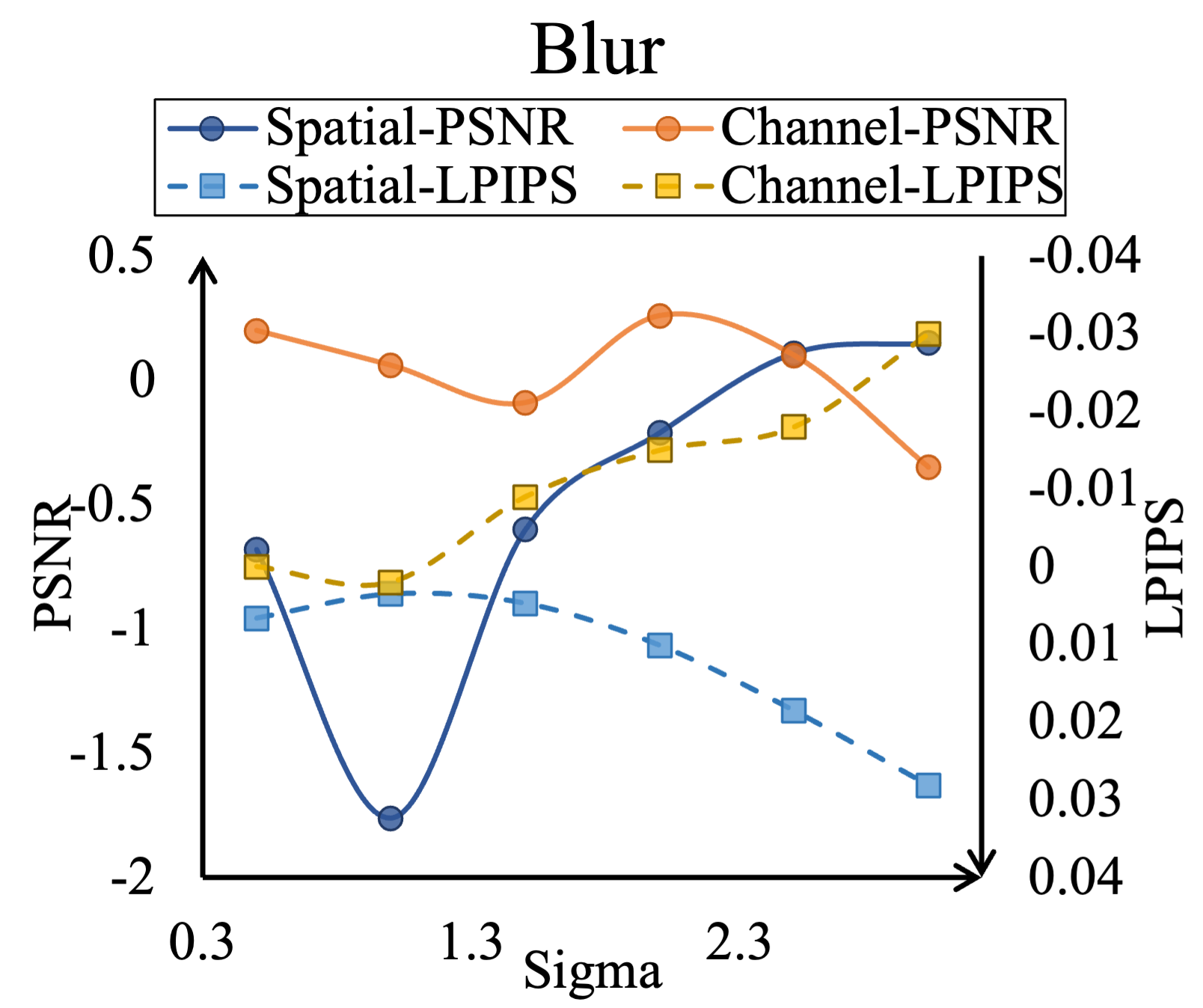}
\end{subfigure}
\begin{subfigure}[t]{0.32\textwidth}
    \centering
    \includegraphics[width=\linewidth]{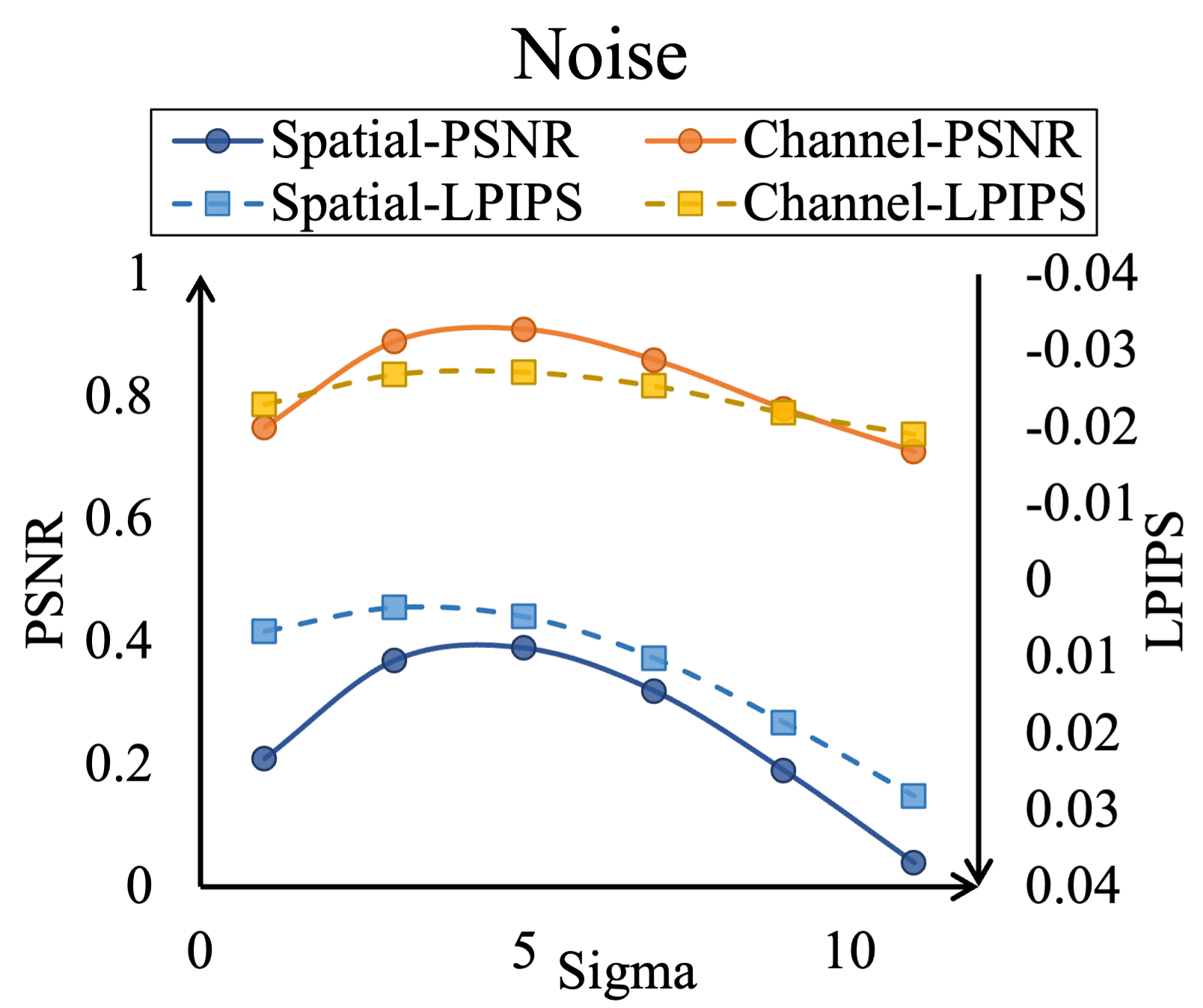}
\end{subfigure}
\begin{subfigure}[t]{0.32\textwidth}
    \centering
    \includegraphics[width=\linewidth]{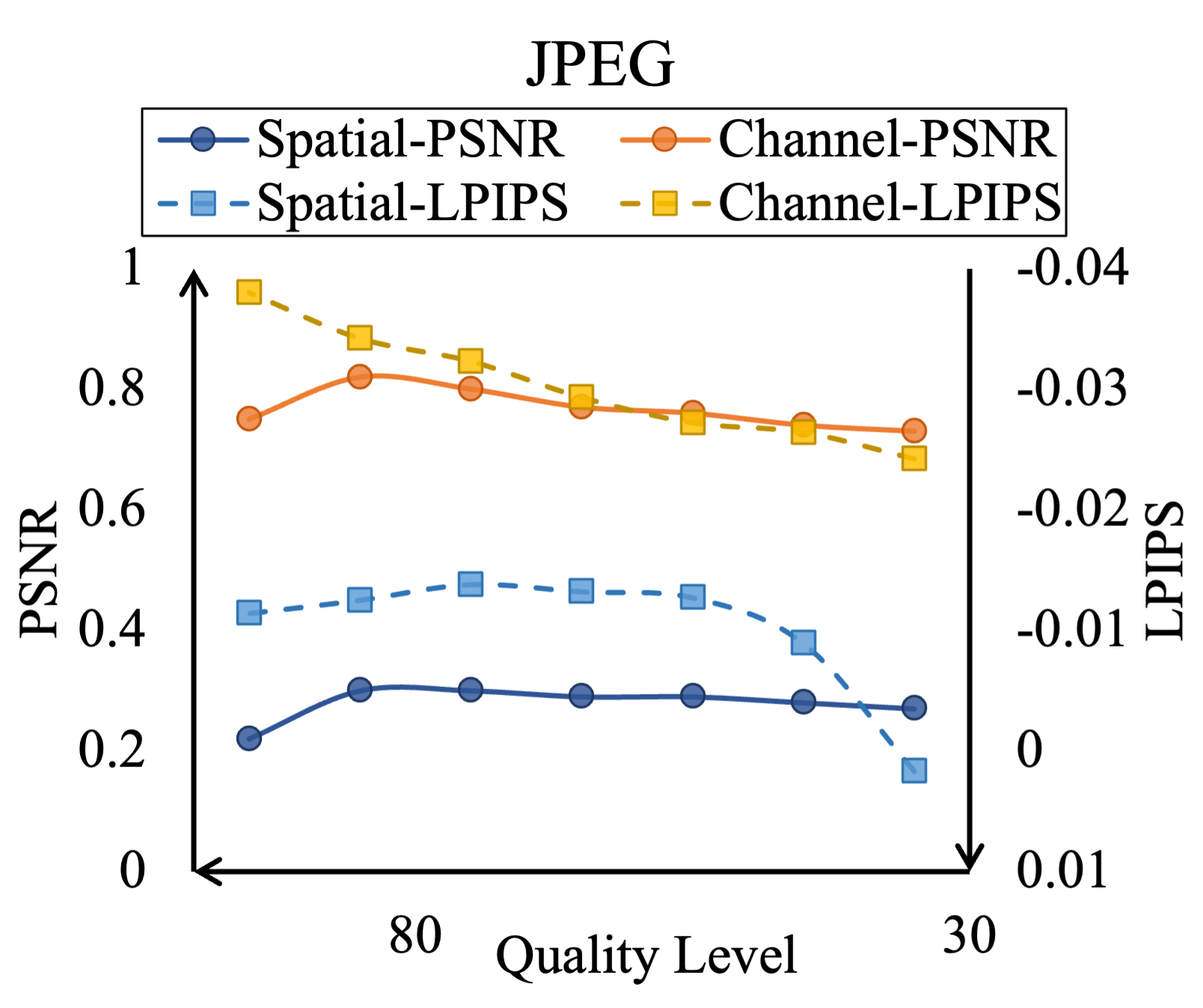}
\end{subfigure}
\caption{Comparison of spatial and channel attention through impact on the performance of real-world VSR model. The Y-axis shows improvements compared to the convolutional baseline. A lower LPIPS score is better. The channel attention module is the best except for the PSNR score of highly blurred inputs.}
\label{fig:sub1_changewithextent}
\end{figure*}

\noindent\textbf{Impact on VSR Models.}
While channel attention output features are less sensitive to query artifacts than spatial attention, how well do they fare for real-world VSR? We experiment by incorporating these attention variants into VSR models. We focus specifically on recurrent pipelines due to their popularity in standard and real-world VSR~\cite{liang2022recurrent, chan2022investigating,xie2022mitigating}.
In recurrent pipelines, 
artifacts in hidden states may propagate over time, get exaggerated, and negatively influence overall performance~\cite{chan2022investigating, xie2022mitigating}. 
Therefore, we explicitly add the attention module between features of the current frame and the propagated hidden state. 
Ideally, the added attention will select relative information from hidden states for the current frame and reduce the propagation of model-produced artifacts.

\cref{subfig:baseline} shows the \textit{baseline} model which concatenates $f_t$, the shallow feature of at time $t$, and $\hat{h}_{t-1}$, the spatially aligned hidden state at time $t-1$. 
We experiment with two variants by using a channel-attention module $\mathrm{G_c}$ or a spatial-attention module $\mathrm{G_s}$ to replace the concatenation in the baseline. 
As shown in \cref{subfig:ch_merge}, the query is generated by $f_{t}$, and the key and value are predicted from $\hat{h}_{t-1}$. The attention output ${O}^{A}_{t}$ is concatenated with $f_{t}$ as the input for $\mathcal{R}$. 

All models are trained on the REDS dataset~\cite{nah2019ntire} with the random degradation pipeline from Real-ESRGAN~\cite{wang2021real}.
We test on REDS4~\cite{nah2019ntire} with different types and extents of degradations, including Gaussian blur, Gaussian noise, and JPEG compression. \cref{fig:sub1_changewithextent} displays changes in PSNR and LPIPS compared with the baseline model. Trained on the same dataset and degradation setting, channel attention between temporal information yields better objective and perceptual reconstruction quality than spatial attention and baseline for noise and JPEG compression inputs. Channel attention still achieves better scores for blurred inputs, except for the PSNR performance of severely blurred inputs.

\subsection{Limitations of Channel Attention}
\label{sec:limitation}
The results in~\cref{sec:investigation} indicate that channel attention is better suited for propagating information over time. However, it also has an inherent flaw -- the correlation among output channels will increase since each channel in the attention output is a weighted summation over channels of the value $V_c$.
We calculate the covariance following VICReg~\cite{bardes2021vicreg} for a quantitative measurement. Given deep feature $z^{n}\in\mathbb{R}^{C\times\!H\times\!W}$, where $n\in\{1,N\}$ is the index of a sample, % from a set of size $N$, 
we reshape $z^{n}$ to $\mathbb{R}^{C\times\!HW}$ and define the covariance matrix over $Z=\{z^{1},...,z^{N}\}$ as:
\begin{equation}
    \textit{Cov}(Z) = \frac{1}{N-1}\sum_{n=1}^{N}(z_n-\bar{z})(z_n-\bar{z})^{T},
\end{equation}
where $\bar{z}=\frac{1}{N}\textstyle \sum_{n=1}^{N} z_n$. We then define the indicator for the covariance matrix as $ac(Z)=\frac{1}{d}\sum_{i\ne\!j}|\textit{Cov}(Z)|_{i,j}$, \ie the average of absolute \textit{off-diagonal} coefficients of $\textit{Cov}(Z)$, where $d$ is the number of off-diagonal coefficients. {Function $ac(Z)$ encodes the covariance among feature channels.}
Taking $O$ in~\cref{fig:query_change} for comparison, $ac(O)$ with $N\!=\!400$ based on channel attention is 0.87 and significantly higher than the input features ($\approx0.15$). Instead, $ac(O)$ of spatial attention remains similar to the input features.
Previous works on representation learning~\cite{bardes2021vicreg} and overfitting~\cite{cogswell2015reducing} propose that the high covariance of feature channels indicates redundancy and tends to hinder informative predictions.

Similarly, we speculate the redundancy effects will hinder the prediction of HR outputs when adopting channel attention in building blocks. For a closer look, we investigate the standard VSR task, which is artifact-free. 
Specifically, we build $M_c$ and $M_s$ by replacing the \textit{Residual Block} in \cref{subfig:baseline} with channel attention blocks~\cite{zamir2022restormer} and spatial attention blocks~\cite{liang2021swinir} 
Models are trained on the REDS dataset without extra degradation. For evaluation, we choose SSIM~\cite{wang2004image}, which focuses on structural information. 
The SSIM score of $M_c$ (0.8338) is lower than $M_s$ (0.8432); $ac(\cdot)$ of the last features before the upsampling module is higher for channel attention than for spatial attention, \ie 0.199 vs. 0.147.

\begin{figure}[!b]
    \centering
    \includegraphics[width=0.6\linewidth]{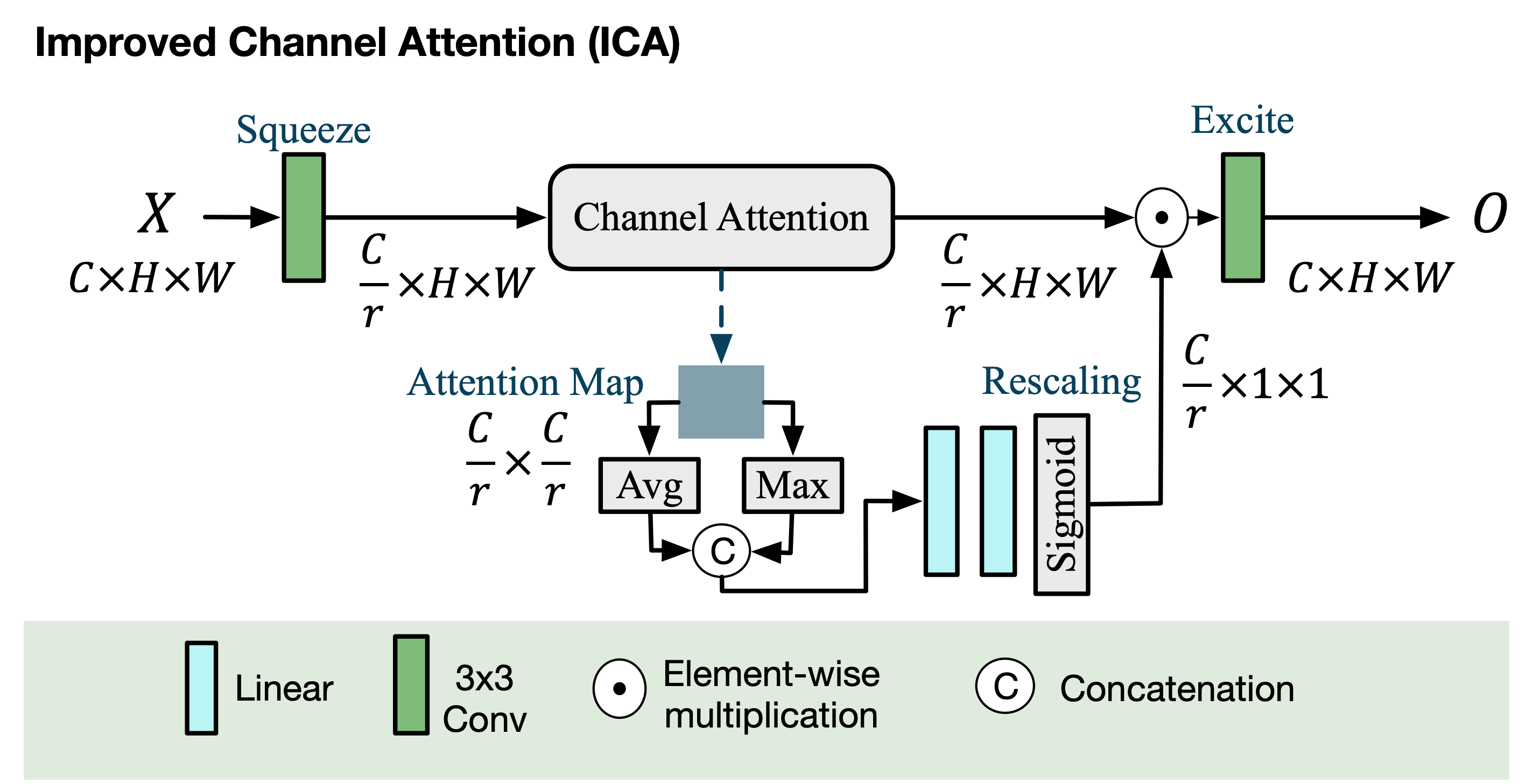}
    \caption{
    Improved Channel Attention Module (ICA), showing self-attention for simplicity. The `squeeze' convolution compresses the number of input feature channels $X\in\mathbb{R}^{C\times\!H\times\!W}$ by ratio $r$. The features are then rescaled by weights predicted from the $\frac{C}{r}\times\frac{C}{r}$ attention map before being expanded by the `excite' convolution back to the original number of input channels.
    } 
    \label{fig:ch_attn}
\end{figure}

We adopt two simple modifications to channel attention and boost informative features to verify the redundancy effect empirically.  \cref{fig:ch_attn} shows the Improved Channel Attention (ICA) module. Our approach features two key steps to enhance the quality. First, we use the squeeze-and-excite mechanism to predict new information. The features are squeezed channel-wise to extract meaningful information and then %using the 'excite' operation to generate new channels 
expanded into new channels based on the attention outputs. Secondly, we rescale channels in attention outputs by scalar weights predicted from the attention map. The attention map measures relationships across channels and encourages the associated convolutions in the 'excite' operation to yield more precise and discriminating features.
Our designs are inspired by the SE Network~\cite{hu2018squeeze} but with two key distinctions. First, we apply the squeeze-and-excite mechanism to generate new information, while SE Network uses it to help with scalar weight prediction. Second, the rescaling weights prediction takes the attention map as a cue rather than the na\"{i}ve pooling results of channels. In ~\cref{sec:method}, we show the efficacy of new designs on the real-world VSR task and compare channel correlation with and without new designs in ablations. 
\begin{figure}[t!]
    \centering
    \includegraphics[width=0.60\linewidth]{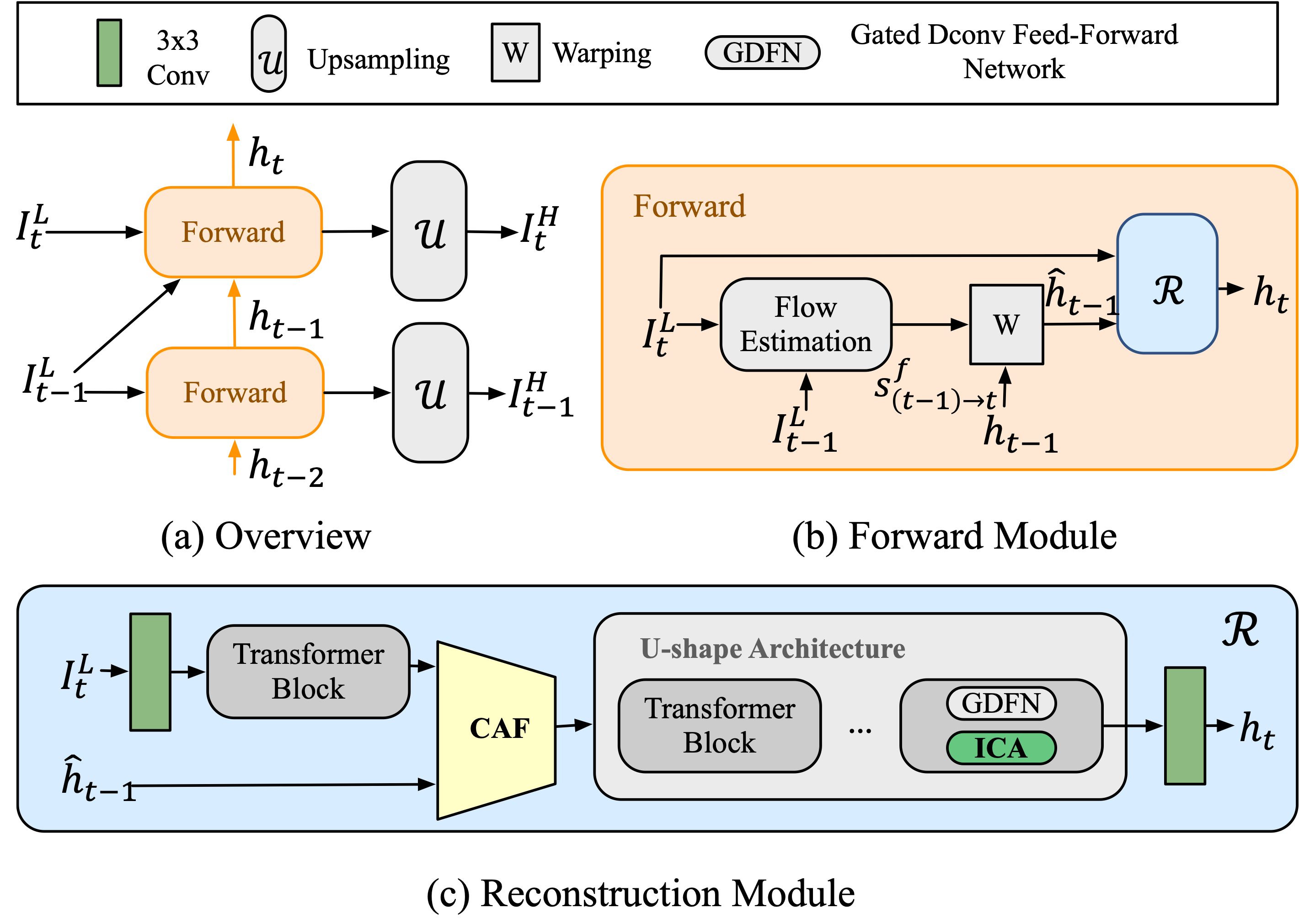}
    \caption{{The framework of RealViformer. (a) Overview of RealViformer, following a unidirectional recurrent framework. The outputs of the Forward module are propagated to the next time step and upsampled by module $\mathcal{U}$ to get HR frames. (b) Explanation of the Forward module in (a), where $W$ denotes the warping function. The reconstruction module $\mathcal{R}$ takes current frame $I^L_t$ and warped hidden state $\hat{h}_{t-1}$ as inputs. (c) Reconstruction module $\mathcal{R}$. The shallow feature of $I^L_t$ and $\hat{h}_{t-1}$ are fused by CAF and then forwarded to Transformer blocks with U-shape connection~\cite{zamir2022restormer}. Module GDFN follows Restormer~\cite{zamir2022restormer}. Details of CAF and ICA modules are stated in \cref{fig:CAF} and \cref{fig:ch_attn}.}}
    \label{fig:archs}
\end{figure}
\section{RealVifromer}
\label{sec:method}
\subsection{Overview}

We apply our findings in \cref{sec:investigation} to the real-world VSR task as further support. To that end, we design RealViformer to incorporate channel attention as the basic processing module and the modification to boost informative features. Our emphasis here is not novel architecture design but a showcase of effectiveness brought by applying our findings in \cref{sec:investigation}.

RealViformer is a recurrent Transformer network with channel attention modules. \cref{fig:archs} shows the model architecture. To reduce the computational cost, we follow a widely used recurrent architecture~\cite{chan2021basicvsr} in a unidirectional setting. The model first estimates optical flow $s^f_{(t-1)\rightarrow t}$ from $I^L_{t-1}$ to $I^L_t$ through Spynet~\cite{ranjan2017optical} and warps previous hidden state $h_{t-1}$ to current time step based on the flow. Frame $I^L_t$ and spatially aligned hidden state $\hat{h}_{t-1}$ are processed in the reconstruction module $\mathcal{R}$. The hidden state is updated with outputs of $\mathcal{R}$ and further processed by the upsampling module $\mathcal{U}$ to output high-resolution frames.

The reconstruction module $\mathcal{R}$ uses channel attention in two ways. 
First, the Channel Attention Fusion (CAF) module fuses the temporal information to limit the produced artifacts in the hidden state. CAF queries the aligned hidden state $\hat{h}_{t-1}$ by the shallow feature $f_t$. Secondly, we take the Improved Channel Attention Module (ICA) in \cref{fig:ch_attn} to build the Transformer blocks for better HR reconstructions. 
% \hl{RMDTA} 

\begin{figure}[!t]
    \centering
    \includegraphics[width=0.70\linewidth]{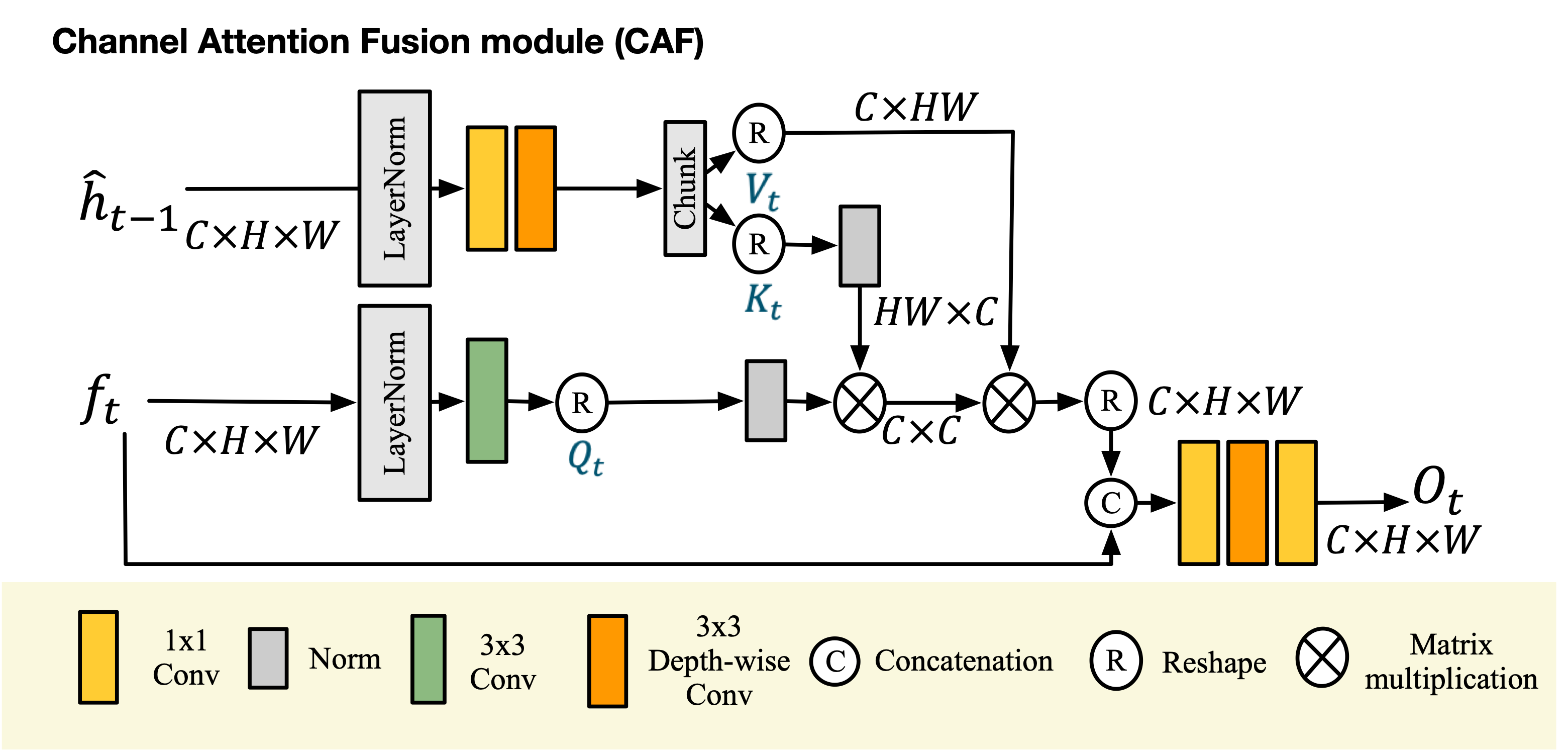}
    \caption{Details of Channel Attention Fusion (CAF) module. CAF gets the query from current frame feature $f_t$ and \{key, value\} from hidden state $\hat{h}_{t-1}$. The attention output is concatenated with $f_t$ to process for module output $O_t$.}
    \label{fig:CAF}
\end{figure}

\subsection{Channel Attention Fusion (CAF) Module }
As explored in \cref{sec:investigation}, adding channel attention promotes the real-world VSR performance compared to spatial attention and the simple concatenation baseline.
Thus, we keep this design in our model and put it as the Channel Attention Fusion (CAF) module for temporal aggregation.
\cref{fig:CAF} show details of how CAF perform channel attention between $f_t$, the feature of the current frame, and $\hat{h}_{t-1}$, the aligned propagated hidden state.
The query ($Q_t$) is generated as $Q_t=K_{3\times\!3}\ast\!\text{LayerNorm}(f_t)$, where $K_{3\times\!3}\ast$ refers to $3\times\!3$ convolution operation. Similarly, we process hidden state $\hat{h}_{t-1}$ by $\Tilde{h}_{t-1}=K^{d}_{3\times\!3}\ast\!K_{1\times\!1}\ast\!\text{LayerNorm}(\hat{h}_{t-1})$, where $K^{d}_{3\times\!3}\ast$ is the $3\times\!3$ depth-wise convolution which double the channel number. 
Chunking $\Tilde{h}_{t-1}$ gives key ($K_t$) and value ($V_t$). 
The calculation of attention map $A_t\in\mathbb{R}^{C\times C}$ follows \cref{eq:attention_c}. The final output $O_t$ is computed as $O_t=K_{1\times\!1}\ast\!K^{d}_{3\times\!3}\ast\!K_{1\times\!1}\ast\!\mathbf{C}[A_tV_t; f_t]$, where $\mathbf{C}[\cdot;\cdot]$ denotes concatenation.

\subsection{Improved Channel Attention (ICA) Module} 
The design of ICA follows \cref{fig:ch_attn} for empirically verifying findings in \cref{sec:limitation}. 
\noindent\textbf{Squeeze and Excite} follows the squeeze-and-excite mechanism in \cref{fig:ch_attn} and helps to predict channels with new information. It squeezes the channels of input feature $X\in\mathbb{R}^{C\times H\times W}$ by factor $r$. The attention map refers to self-attention on $X$ and of size $\mathbb{R}^{\frac{C}{r} \times \frac{C}{r}}$. Excitation layers expand outputs back to size ${\mathbb{R}^{C \times H\times W}}$.

\noindent\textbf{Correlation-based channel weighting} weights the output channel based on scalars predicted from the attention map. This design emphasizes channels for predicting more discriminative features in `excite' operation.
The attention map $A_r \in \mathbb{R}^{\frac{C}{r} \times \frac{C}{r}}$ is taken for calculating the average and max values along the rows. The average and max values are combined and mapped to weights of size $\mathbb{R}^{\frac{C}{r} \times 1}$ through linear layers and a sigmoid function.

\subsection{Implementation Details}
\label{sec:details}
\noindent\textbf{Model details.} RealViformer uses SPyNet~\cite{ranjan2017optical} for flow estimation. After the CAF module, the reconstruction module $\mathcal{R}$ has a three-level encoder-decoder architecture. From level 1 to level 3, there are [2,3,4] transformer blocks with [48,96,192] channels. There are [1,2,4] attention heads in the ICA, all with a squeeze factor of 4. Supplementary B.1 gives the detailed architecture of $\mathcal{R}$. 

\noindent\textbf{Training details.}
We train using the REDS dataset~\cite{nah2019ntire} and follow RealBasicVSR~\cite{chan2022investigating} in applying random combinations of blur, noise, JPEG compression, and video compression for synthesizing input degradations. We load 15 frames as an input sequence. The spatial size of inputs is cropped to $64\times 64$, and the batch size is 16. We use a pre-trained flow estimation model SPyNet, the parameter of which is fixed for the first 5K iterations and tuned with other modules later. 

Following RealBasicVSR~\cite{chan2022investigating}, we perform two-stage training. The first stage trains the model with a Charbonnier loss~\cite{lai2018fast} and SSIM~\cite{wang2004image} loss for 300K iterations.
In the second stage, the model is trained for another 130K iterations with the Charbonnier loss, SSIM loss, perceptual loss~\cite{johnson2016perceptual} and GAN loss~\cite{goodfellow2020generative} together, weighted by 1, 0.001, 1, and 0.005, respectively. The implementations of perceptual loss, GAN loss, and discriminator follow RealBasicVSR~\cite{chan2022investigating}. 
We implement all experiments on 4 Quadro RTX 8000 GPUs with PyTorch. Other details of the training settings are in Supplementary B.2.
\begin{table*}[h]
    \centering
    \caption{Quantitative comparisons with existing methods with \textbf{best} and \underline{second-best} results. Our method achieves the best ILNIQE and NRQM scores over VideoLQ and RealVSR datasets and the best PSNR, SSIM, and LPIPS over synthetic datasets REDS4 and UDM10 with relatively few parameters and the lowest run-time.}
    \resizebox{\textwidth}{!}{
    \begin{tabular}{c c|c|c|c|c|c|c|c|c}
              & &RealSR & DAN & RealVSR & DBVSR & BSRGAN & Real-ESRGAN & RealBasicVSR &Ours \\
        \hline
         \multicolumn{2}{c|}{Params (M)} &16.7& \underline{4.3} &\textbf{2.7} & 25.5& 16.7& 16.7&6.3&5.3\\
         \multicolumn{2}{c|}{Runtime (ms)}&180 &250 &772 &- &180 &196 &\underline{73} &\textbf{49}\\
         \hline
         \multirow{2}{*}{VideoLQ}&ILNIQE$\downarrow$&26.63 &28.28 &31.94 &27.85 &27.49 &27.97 &\underline{25.98} &\textbf{25.94} \\
         &NRQM$\uparrow$&6.054 &3.742 &3.460 &3.851 &6.156 &6.057 & \underline{6.306} &\textbf{6.338} \\
         \hline
         \multirow{2}{*}{RealVSR}&ILNIQE$\downarrow$& 32.81 & 32.29 &34.39 & - & 32.65 & 31.93 & \underline{30.37} &\textbf{28.61} \\
         &NRQM$\uparrow$& 5.610 & 3.523 &3.795  & - & 6.152 & 6.245 & \underline{6.582} &\textbf{6.588} \\
         \hline
         \multirow{4}{*}{REDS4}&PSNR$\uparrow$&22.02 &22.67 & 18.30&22.35 &22.94 &21.56 &\underline{23.09} & \textbf{23.34}\\
         &SSIM$\uparrow$&0.5097  &0.5571 &0.4900 &0.5530 &0.5750 &0.5556 &\underline{0.6076} &\textbf{0.6079} \\
         &LPIPS$\downarrow$&0.5991  &0.6315 &0.7240 &0.6211 &0.3766 &0.3533 &\underline{0.2991} &\textbf{0.2877} \\
         \hline
         \multirow{4}{*}{UDM10}&PSNR$\uparrow$&25.37  &25.90 &23.35 &25.08 &\underline{25.97} &24.96 &25.96 & \textbf{26.42}\\
         &SSIM$\uparrow$&0.6658  &0.7229 &0.7115 &0.7112 &\underline{0.7568} & 0.7432 &0.7491 & \textbf{0.7609}\\
         &LPIPS$\downarrow$&0.4811  &0.4781 &0.4761 &0.4756 &0.3388 &0.3395 &\underline{0.3209} & \textbf{0.3063}\\
         % &NRQM$\uparrow$&  & & & & & & & 
    \end{tabular}}
    \label{tab:quantitative resuts}
\end{table*}
\subsection{Experimental Results}
We compare our model on four datasets, VideoLQ~\cite{chan2022investigating}, RealVSR~\cite{yang2021real}, REDS4~\cite{nah2019ntire} and UDM10~\cite{yi2019progressive}, with RealBasicVSR~\cite{chan2022investigating}, Real-ESRGAN~\cite{wang2021real}, BSRGAN~\cite{zhang2021designing}, DBVSR~\cite{pan2021deep}, RealVSR~\cite{yang2021real}, DAN~\cite{huang2020unfolding}, and RealSR~\cite{ji2020real}. VideoLQ and RealVSR are collected from real-world scenarios.
VideoLQ is an unpaired dataset. RealVSR has same-size paired low-quality (LQ) and high-quality (HQ) frames. Our model super-resolves the %spatial size of 
input frames spatially. Although downsampling LQ enables paired data, it alters the original degradation. Thus, we test on the original LQ and do not use the HQ for evaluation. REDS4 and UDM10 have ground-truth images, and we synthesize low-quality inputs through the same pipeline during training.
All tested recurrent-based VSR models load the half sequence for videos in RealVSR and the whole sequence for others each time. The quantitative and qualitative results are discussed below.

\noindent\textbf{Quantitative Results.} For evaluation without reference, we apply the ILNIQE~\cite{zhang2015feature}, the improved version of NIQE~\cite{mittal2012making}, and NRQM~\cite{ma2017learning} metrics based on each output sequence's first, middle, and last frames in RGB format~\cite{chan2022investigating}. These metrics appear less biased towards oversharpened features; details are given in Supplementary B.3. For evaluation on REDS4 and UDM10, we report more reliable metrics, PSNR, SSIM, and LPIPS~\cite{zhang2018unreasonable}. We collect released models of all compared methods and generate sequences. As shown in \cref{tab:quantitative resuts}, RealViformer performs better than other methods with smaller parameters than the most competitive RealBasicVSR~\cite{chan2022investigating} and the shortest runtime.

\begin{figure}[h]
    \centering
    \begin{subfigure}[t]{0.19\textwidth}
    \centering
        \includegraphics[width=\linewidth]{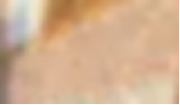}
    \end{subfigure}
    \begin{subfigure}[t]{0.19\textwidth}
    \centering
        \includegraphics[width=\linewidth]{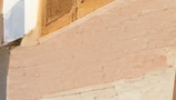}
        % \subcaption{ NLSA}}
    \end{subfigure}
    \begin{subfigure}[t]{0.19\textwidth}
    \centering
        \includegraphics[width=\linewidth]{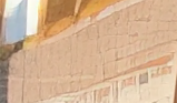}
        % \subcaption{ RankSRGAN}}
    \end{subfigure}
    \begin{subfigure}[t]{0.19\textwidth}
    \centering
        \includegraphics[width=\linewidth]{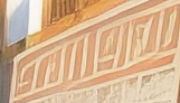}
        % \subcaption{ WDST}}
    \end{subfigure}
    \begin{subfigure}[t]{0.19\textwidth}
    \centering
        \includegraphics[width=\linewidth]{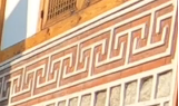}
        % \subcaption{ Our Result}}
    \end{subfigure}

    \begin{subfigure}[t]{0.19\textwidth}
    \centering
        \includegraphics[width=\linewidth]{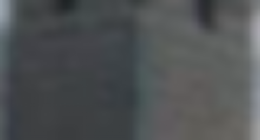}
        % \subcaption{Input}
    \end{subfigure}
    \begin{subfigure}[t]{0.19\textwidth}
    \centering
        \includegraphics[width=\linewidth]{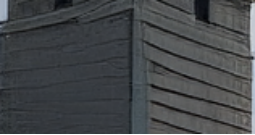}
        % \subcaption{RE}
    \end{subfigure}
    \begin{subfigure}[t]{0.19\textwidth}
    \centering
        \includegraphics[width=\linewidth]{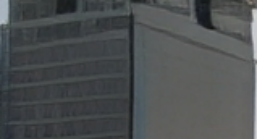}
        % \subcaption{RB}
    \end{subfigure}
    \begin{subfigure}[t]{0.19\textwidth}
    \centering
        \includegraphics[width=\linewidth]{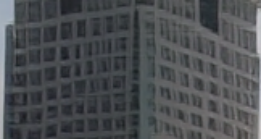}
        % \subcaption{Ours}
    \end{subfigure}
    \begin{subfigure}[t]{0.19\textwidth}
    \centering
        \includegraphics[width=\linewidth]{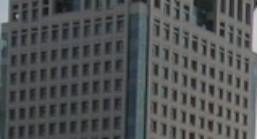}
        % \subcaption{Ground Truth}
    \end{subfigure}
    
    \begin{subfigure}[t]{0.19\textwidth}
    \centering
        \includegraphics[height=0.55\linewidth,width=\linewidth]{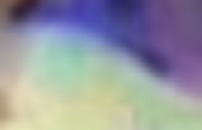}
        \subcaption{Input}
    \end{subfigure}
    \begin{subfigure}[t]{0.19\textwidth}
        \centering
        \includegraphics[height=0.55\linewidth,width=\linewidth]{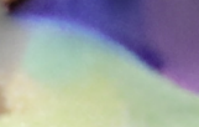}
        \subcaption{RE}
    \end{subfigure}
    \begin{subfigure}[t]{0.19\textwidth}
        \centering
        \includegraphics[height=0.55\linewidth,width=\linewidth]{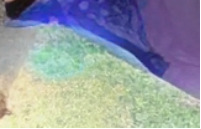}
        \subcaption{RB}
    \end{subfigure}
    \begin{subfigure}[t]{0.19\textwidth}
    \centering
        \includegraphics[height=0.55\linewidth,width=\linewidth]{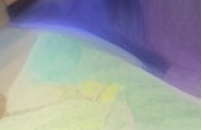}
        \subcaption{Ours}
    \end{subfigure}
    \begin{subfigure}[t]{0.19\textwidth}
    \centering
        \includegraphics[height=0.55\linewidth,width=\linewidth]{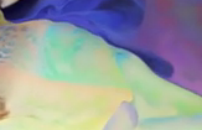}
        \subcaption{GT}
    \end{subfigure}
    \caption{{Qualitative comparisons on synthetic datasets. Our method produces clearer than RealESRGAN (RE)~\cite{wang2021real} and RealBasicVSR (RB)~\cite{chan2022investigating} for very hard inputs.}}
    \label{fig:qualitative_cmp_syn}
\end{figure}
\begin{figure}[h]
    \centering
    \begin{subfigure}[t]{0.24\textwidth}
    \centering
        \includegraphics[width=\linewidth]{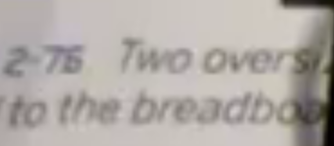}
        % \subcaption{ Ground Truth}}
    \end{subfigure}
    \begin{subfigure}[t]{0.24\textwidth}
    \centering
        \includegraphics[width=\linewidth]{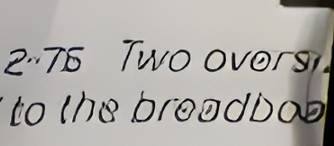}
        % \subcaption{ NLSA}}
    \end{subfigure}
    \begin{subfigure}[t]{0.24\textwidth}
    \centering
        \includegraphics[width=\linewidth]{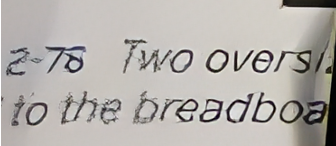}
        % \subcaption{ RankSRGAN}}
    \end{subfigure}
    \begin{subfigure}[t]{0.24\textwidth}
    \centering
        \includegraphics[width=\linewidth]{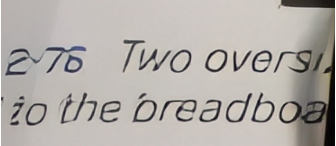}
    \end{subfigure}

    \begin{subfigure}[t]{0.24\textwidth}
    \centering
        \includegraphics[width=\linewidth]{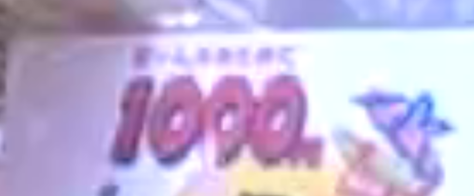}
        % \subcaption{ Ground Truth}}
    \end{subfigure}
    \begin{subfigure}[t]{0.24\textwidth}
    \centering
        \includegraphics[width=\linewidth]{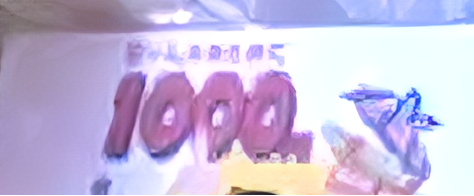}
        % \subcaption{ NLSA}}
    \end{subfigure}
    \begin{subfigure}[t]{0.24\textwidth}
    \centering
        \includegraphics[width=\linewidth]{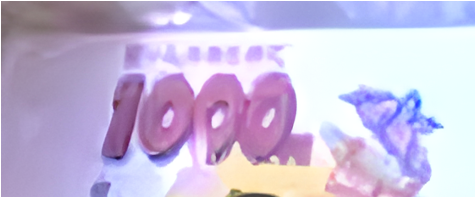}
        % \subcaption{ RankSRGAN}}
    \end{subfigure}
    \begin{subfigure}[t]{0.24\textwidth}
    \centering
        \includegraphics[width=\linewidth]{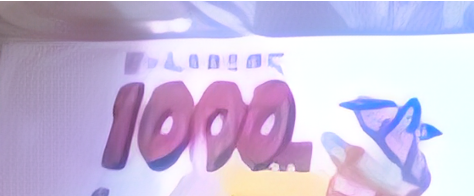}
    \end{subfigure}
    
    \begin{subfigure}[t]{0.24\textwidth}
    \centering
        \includegraphics[height=0.45\linewidth,width=\linewidth]{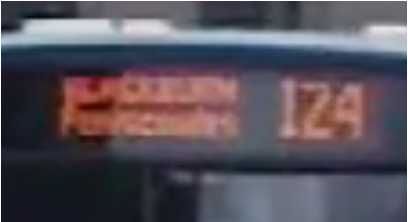}
        \subcaption{Input}
    \end{subfigure}
    \begin{subfigure}[t]{0.24\textwidth}
        \centering
        \includegraphics[height=0.45\linewidth,width=\linewidth]{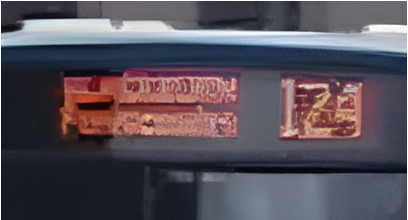}
        \subcaption{RE}
    \end{subfigure}
    \begin{subfigure}[t]{0.24\textwidth}
    \centering
        \includegraphics[height=0.45\linewidth,width=\linewidth]{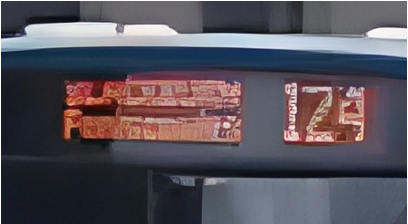}
        \subcaption{RB}
    \end{subfigure}
    \begin{subfigure}[t]{0.24\textwidth}
    \centering
        \includegraphics[height=0.45\linewidth,width=\linewidth]{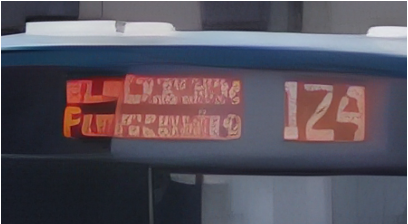}
        \subcaption{Ours}
    \end{subfigure}
    \caption{{Qualitative comparisons on real-world datasets. Our method produces less high-frequency artifacts and overshoot effects than RealESRGAN~\cite{wang2021real} and RealBasicVSR~\cite{chan2022investigating}}.}
    \label{fig:qualitative_cmp_real}
\end{figure}

\noindent\textbf{Qualitative Results.} We show qualitative comparisons on synthetic (see \cref{fig:qualitative_cmp_syn}) and real-world (see \cref{fig:qualitative_cmp_real}) datasets. Compared to the listed methods, RealViformer generates clear structures with much fewer high-frequency artifacts. More visual comparisons are in Supplementary B.5.

\begin{figure}[h]
\begin{minipage}[]{0.40\textwidth}
    \centering
\includegraphics[width=0.90\textwidth]{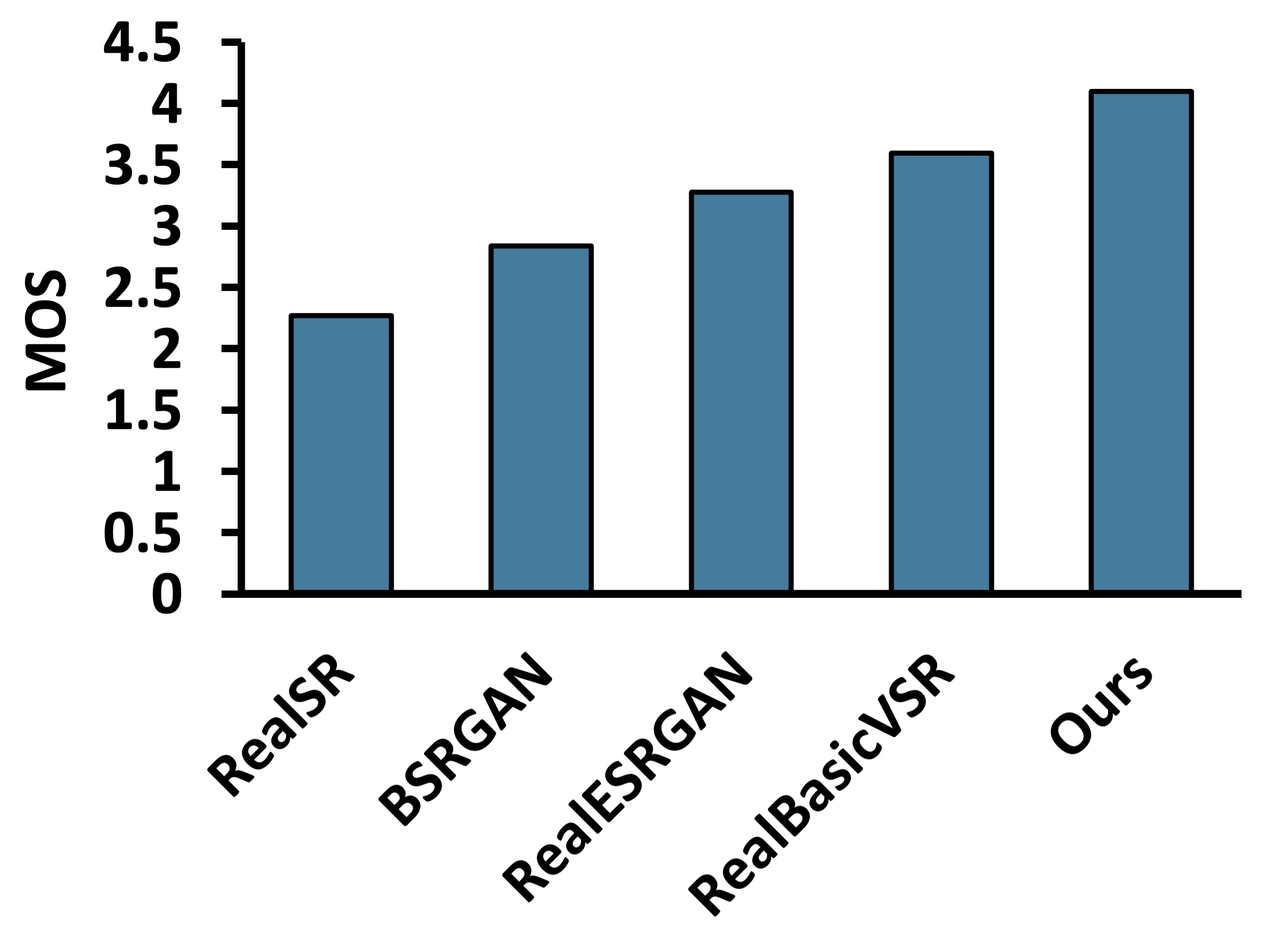}
    \caption{User study results from 30 evaluators on 85 frames. Our method achieves the best among all five methods regrading mean option scores (MOS).}
    \label{fig:userstudy}
\end{minipage}
\hfill
\begin{minipage}{0.58\textwidth}
\centering
    \captionof{table}{Ablations of CAF and ICA modules. Channel-attention baseline (ch-baseline) performs better than the spatial-attention baseline (sp-baseline). The CAF module improves the performances for both datasets, and ICA further improves the performances to the state-of-the-art.}
    \begin{tabular}{c|c c|c c}
         \multirow{2}{*}{Method}&\multirow{2}{*}{CAF} & \multirow{2}{*}{ICA} & VideoLQ & UDM10  \\
         & & & NRQM$\uparrow$ & LPIPS$\downarrow$\\
         \hline
         $\text{Sp-baseline}$ &  -  &  -   & 6.061   & 0.3482  \\
         $\text{Ch-baseline}$& \XSolidBrush & \XSolidBrush  &  6.181 & 0.3085\\
        $\text{RealViformer}^{-}$&\Checkmark & \XSolidBrush  &6.196 & 0.2933\\
        RealViformer&\Checkmark & \Checkmark & 6.338 & 0.2877\\
    \end{tabular}
    \label{tab:ablation}
\end{minipage}
\end{figure}

\noindent\textbf{User Study.} We also performed a user study and asked 30 evaluators on Amazon MTurk to score reconstructions for 85 frames sampled from datasets VideoLQ~\cite{chan2022investigating}, RealVSR~\cite{yang2021real}, REDS4~\cite{nah2019ntire} and UDM10~\cite{yi2019progressive}. Each worker saw five HR results of the same frame and rated them based on the visual quality, from 1 (the worst) to 5 (the best); as shown in \cref{fig:userstudy}, our method surpasses BSRGAN~\cite{zhang2021designing}, Real-ESRGAN~\cite{wang2021real}, RealSR~\cite{ji2020real}, and RealBasicVSR~\cite{chan2022investigating}.

\begin{figure}[h]
\centering
\begin{minipage}[h]{0.65\textwidth}
\centering
\begin{subfigure}[t]{\textwidth}
    \centering
        \includegraphics[width=\linewidth]{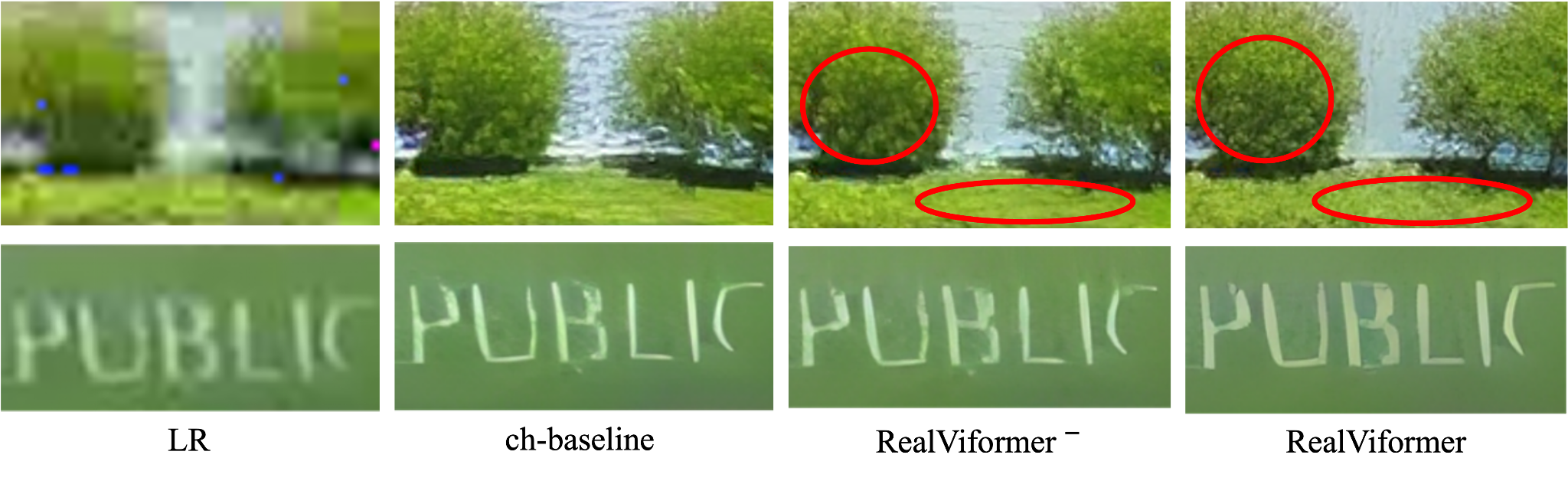}
        \subcaption{Visual comparison}
    \label{fig:abla_vis}
\end{subfigure}
\end{minipage}
\begin{minipage}{0.30\textwidth}
\begin{subfigure}[t]{\textwidth}
    \centering
        \includegraphics[width=0.85\linewidth]{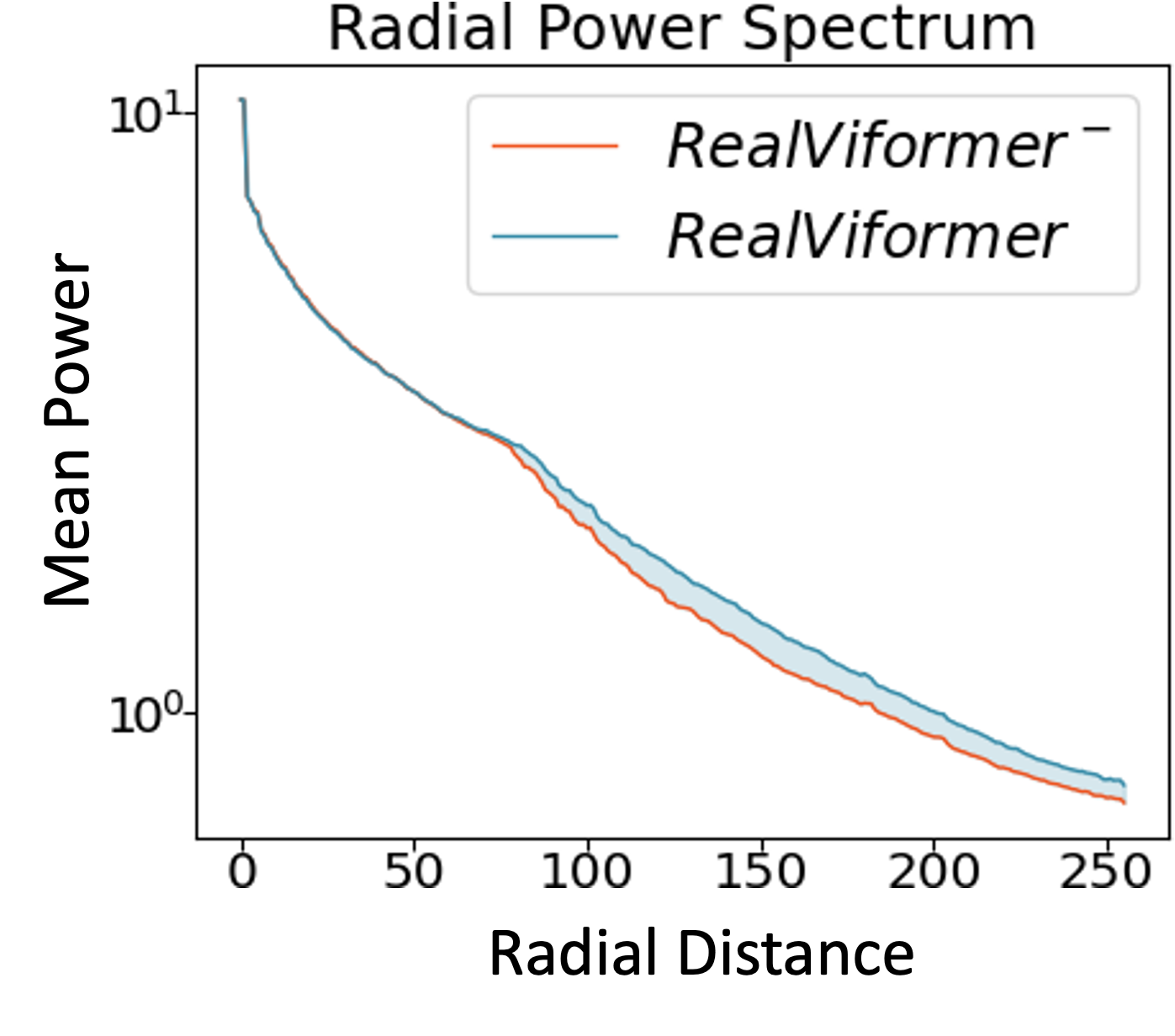}
        \subcaption{RPS}
    \label{fig:abla_rps}
\end{subfigure}
\end{minipage}
\caption{(a) Visual comparison between RealViformer and its ablations. Red circles highlight the improved details. (b) Radial Power Spectrum (RPS) of model predictions. Using ICA improves the power of high-frequency components (blue region).}
\end{figure}

\noindent\textbf{Ablation Studies.} We conduct ablations to validate the advantage of channel attention and the efficacy of the CAF and ICA. We build a spatial-attention baseline (sp-baseline) by replacing the reconstruction module of BasicVSR\cite{chan2021basicvsr} with SwinIR~\cite{liang2021swinir}. Channel-attention baseline (ch-baseline) is built with the same overall architecture as RealViformer but replaces CAF with simple concatenation and substitutes ICA with the channel attention block in Restormer~\cite{zamir2022restormer}. $\text{RealViformer}^{-}$ applies CAF on ch-baseline. 
All models are trained with the same settings in \cref{sec:details}. We report NRQM scores for the VideoLQ~\cite{chan2022investigating} dataset and LPIPS scores for the UDM10~\cite{yi2019progressive} da aset. As shown in~\cref{tab:ablation}, using the original channel attention module, $\text{ch-baseline}$ already yields better performances than sp-baseline.
The CAF and ICA modules further improve RealViformer to state-of-the-art performance with channel correlation of propagated information decreasing from 0.436 to 0.422. \cref{fig:abla_vis} visually compares RealViformer with its ablations. Applying CAF reduces artifacts, while ICA provides further improvements. \cref{fig:abla_rps} supplements the Radial Power Spectrum of model predictions. The blur region shows ICA increases the power in the high-frequency region.

\section{Conclusion}
This paper proposes a real-world VSR model, RealViformer, based on findings from investigating channel and spatial attention in a real-world setting. Explorations reveal that channel attention is less sensitive to the artifacts in query and better serves as a temporal aggregation module to limit model-produced artifacts in hidden states. Additionally, we observe the higher covariance of channel attention outputs and propose the Improved Channel Attention (ICA) Module with a squeeze-and-excite and a covariance-based rescaling mechanism. Based on our findings, we build RealViformer, a channel-attention-based recurrent model for real-world VSR. We propose the CAF module to limit artifact propagation and use the ICA module to achieve better reconstructions. RealViformer performs state-of-the-art on two real-world video datasets with fewer parameters and shorter runtime. On the other hand, we value our findings \wrt comparisons between channel and spatial attention and exploration of covariance in channel attention as inspiration for further real-world VSR research.

% \clearpage  % TODO REVIEW/FINAL: This \clearpage needs to be removed from both review and camera-ready versions.
\subsubsection{Acknowledgement}
This research is supported by the National Research Foundation, Singapore under its NRF Fellowship for AI (NRF-NRFFAI1-2019-0001). Any opinions, findings and conclusions or recommendations expressed in this material are those of the author(s) and do not reflect the views of National Research Foundation, Singapore.

\bibliographystyle{splncs04}
\bibliography{main}
\end{document}